\documentclass[preprint,12pt]{elsarticle}

\usepackage{epsfig}
\usepackage{subcaption}
\usepackage{calc}
\usepackage{amssymb}
\usepackage{amstext}
\usepackage{amsmath}
\usepackage{amsthm}
\usepackage{multicol}
\usepackage{pslatex}
\usepackage{apalike}
\usepackage{xcolor}
\usepackage{hyperref}
\usepackage[bottom]{footmisc}
\usepackage{algpseudocode}
\usepackage{multirow}
\usepackage{comment}

\begin{document}
\begin{frontmatter}

\title{Optimizing 2D+1 Packing in Constrained Environments Using Deep Reinforcement Learning}

\author[unifesp]{Victor Ulisses Pugliese}
\author[embraer]{Os\'{e}ias F. de A. Ferreira}
\author[ist]{Fabio A. Faria}

\begin{abstract} 
This paper proposes a novel approach based on deep reinforcement learning (DRL) for the 2D+1 packing problem with spatial constraints. This problem is an extension of the traditional 2D packing problem, incorporating an additional constraint on the height dimension. Therefore, a simulator using the OpenAI Gym framework has been developed to efficiently simulate the packing of rectangular pieces onto two boards with height constraints. Furthermore, the simulator supports multidiscrete actions, enabling the selection of a position on either board and the type of piece to place. Finally, two DRL-based methods (Proximal Policy Optimization -- PPO and the Advantage Actor-Critic -- A2C) have been employed to learn a packing strategy and demonstrate its performance compared to a well-known heuristic baseline (MaxRect-BL). In the experiments carried out, the PPO-based approach proved to be a good solution for solving complex packaging problems and highlighted its potential to optimize resource utilization in various industrial applications, such as the manufacturing of aerospace composites.

\end{abstract}

\end{frontmatter}

\section{Introduction}
Manufacturing has undergone significant changes in recent decades, primarily driven by market trends that encourage companies to transition from traditional mass production lines to more dynamic and flexible manufacturing systems, essential for competitiveness in the global market. This shift, known as smart manufacturing, is currently reinventing itself through advances in Digital Transformation, Internet of Things (IoT), and Artificial Intelligence (AI) ~\cite{alemao2021smart}, ~\cite{xia2021digital}, and ~\cite{ramezankhani2021making}.

Consequently, various approaches to manufacturing scheduling have been studied and implemented to optimize production and resource allocation. Despite these efforts, most scheduling uses manual methods or basic software, resulting in limited improvements in system performance. Historically, the production lines produced many of the same products, always following the same process. However, this is not the case for Smart Manufacturing ~\cite{alemao2021smart}.

Aerospace manufacturing, particularly using composite materials, presents a complex scheduling challenge characterized by high demand variability, extended lead times, and the integration of diverse suppliers and work practices. Although composites offer advantages such as superior strength, corrosion resistance, and efficient forming, their higher cost than traditional metallic materials requires careful optimization \cite{xie2020two} and~\cite{azami2018scheduling}. The manufacturing process typically involves two primary stages: layup and curing \cite{azami2016scheduling}. Autoclave packing, a critical aspect of the curing process, involves meticulous placement of composite parts within the autoclave to achieve desired product properties ~\cite{haskilic2023real} and ~\cite{elkington2015hand}. This intricate task, involving manual positioning, presents a unique optimization problem that surpasses the classical packing problem due to additional constraints and resource management requirements \cite{collart2015application}. 

Certain constraints can be relaxed to simplify the optimization process. For instance, since composite materials cannot be stacked within an autoclave, the placement strategy can focus on the width and length of the parts. Additionally, the height of each part must be verified to ensure it does not exceed the capacity of the tooling cart.

The introduction of Reinforcement Learning (RL) methods to solve packing problems has shown promising results in the literature. For instance, ~\cite{kundu2019deep} employed RL to take an image as input and predict the pixel position of the next box, while~\cite{li2022one} explored RL in 2D and 3D environments. Furthermore, combining heuristics with RL, as in ~\cite{fang2023deep}, has proven to be effective, and RL has also been applied to several other types of problem, as discussed in \cite{wang2022self}. One of the advantages of RL is that it does not require an explicit model of the environment; the agent learns to make decisions by observing the rewards of its actions from a state, as described in \cite{sutton2018reinforcement}, and continuously adapts to its environment through exploration and exploitation. This makes RL particularly suitable for sequential decision-making in games, robotics, control systems, and scheduling problems~\cite{cheng2021heuristic}.

Our approach distinguishes itself by relying solely on RL methods, using actor-critic to explore and exploit. This contrasts with other packing studies that frequently incorporate heuristics to guide or direct the RL algorithm, thereby limiting its scope and creativity. To our knowledge, no scientific study has ever addressed this topic in the literature. Therefore, this paper aims to apply  Reinforcement Learning methods to address a 2D+1 packing problem with spatial constraints. This problem is an extension of the traditional 2D packing problem, incorporating an additional constraint on the height dimension. We also compare the PPO and A2C as the unique methods that support multi-discrete action spaces. This research, inspired by the challenges of aerospace composite manufacturing, has potential applications in many industry sectors, including the packing of components in vehicles, organizing parts in boxes or pallets for transport and storage, arranging products in-store displays, and similar optimization tasks across different sectors.

\section{Background}
This section briefly describes the types of packing problem and the deep reinforcement learning (DRL) methods used in this paper.

\subsection{Packing}\label{subsec:Empacotamento}
The packing problem is a classic challenge in combinatorial optimization that has been extensively studied for decades by researchers in operations research and computer science, as noted in~\cite{li2022one}. 

The primary objective is to allocate objects within containers, minimizing wasted space efficiently. The problem can work with regular \cite{kundu2019deep} and ~\cite{zhao2022reinforcement} or irregular shapes \cite{crescitelli2023deep}, often explored in streaming/online or batching/offline approaches. 

Several works based on heuristic approaches have been proposed for solving packing problems as described in \cite{oliveira2016survey}, such as the {Maximum Rectangles - Bottom-Left (Max Rects-BL)}, {Best-Fit Decreasing Height (BFDH)}, and Next-Fit Decreasing Height (NFDH). Max Rects-BL approach places the largest rectangle in the nearest available bottom-left corner of a 2D space~\cite{fang2023deep}.  {BFDH sorts items by descending height and then attempts to place each item, left-justified, on the existing level with the minimum remaining horizontal space \cite{seizinger2018two}. In the NFDH approach, it first arranges the pieces in descending order of heigthen places each piece on the current level, starting from the left side, as long as there is enough space; otherwise, it starts a new level \cite{oliveira2016survey}.

\subsection{Deep Reinforcement Learning}\label{subsec:AprendizadoReforço}
Deep Reinforcement Learning (DRL) addresses the challenge of autonomously learning optimal decisions over time. Although it employs well-established supervised learning methods, such as deep neural networks for function approximation, stochastic gradient descent (SGD), and backpropagation, RL applies these techniques differently, without a supervisor, using a reward signal and delayed feedback. In this context, an RL agent receives dynamic states from an environment and takes actions to maximize rewards through trial-and-error interactions ~\cite{kaelbling1996reinforcement}.

The agent and the environment interact in a sequence at each discrete time step, $t = 0, 1, 2, 3, \cdots$. At each time step $t$, the agent receives a representation of the environment's state $s_{t} \in S$, where $S$ is the set of possible states, and selects an action $a_{t} \in A(s_{t})$, where $A(s_{t})$ is the set of actions available in the state $s_{t}$. At time step $t + 1$, as a consequence of its actions, the agent receives a numerical reward $r_{t+1} \in R$ and transitions to a new state $s_{t+1}$~\cite{sutton2018reinforcement}.

During each iteration, the agent implements a mapping from states to the probabilities of each possible action. This mapping, known as the agent's policy, is denoted as $\pi_{t}$, where $\pi_{t}(s, a)$ represents the probability that $a_{t} = a$ given $s_{t} = s$. Reinforcement learning methods specify how the agent updates its policy based on experience, intending to maximize the cumulative reward over the long term, according ~\cite{sutton2018reinforcement}.

\subsubsection{Proximal Policy Optimization (PPO)} PPO employs the actor-critic method and trains on-policy, meaning it samples actions based on the most recent policy iteration ~\cite{schulman2017proximal}. In this framework, two neural networks typically serve as the ``actor" and ``critic." The ``actor" learns the policy, while the ``critic" estimates the value function or the advantage, which is used to train the ``actor".

The training process involves calculating future rewards and advantage estimates to refine the policy and adjust the value function. Both the policy and value function are optimized using stochastic gradient descent algorithms, as described in \cite{ppokeras}.

The degree of randomness in action selection depends on the initial conditions and the training procedure. Typically, as training progresses, the policy becomes less random due to updates that encourage the exploration of previously discovered rewards \cite{saenzevaluating}.

\subsubsection{Advantage Actor-Critic (A2C)}
A2C, often perceived as a distinct algorithm, is revealed in ``A2C is a special case of PPO" as a specific configuration of Proximal Policy Optimization (PPO) operating within the actor-critic approach. A2C shares similarities with PPO in employing separate neural networks for policy selection (actor) and value estimation (critic). Its core objective aligns with PPO when the latter's update epochs are set to 1, effectively removing the clipping mechanism and streamlining the learning process \cite{huang2022a2c}.

A2C is a synchronous adaptation of the Asynchronous Actor-Critic (A3C) policy gradient approach. It operates deterministically, waiting for every actor to complete its experience segment before initiating updates, averaging across all actors. This strategy improves GPU utilization by accommodating larger batch sizes \cite{mnih2016asynchronous}.

\section{Related Works}
The field of 2D regular packing problems has seen significant progress in recent years, with various methods proposed to optimize space utilization and minimize waste, using Reinforcement Learning. This review connects several key research papers, highlighting the diverse strategies to tackle these challenges.

In online 2D bin packing, where items are placed sequentially without prior knowledge of future inputs, \cite{kundu2019deep} propose a variation of DQN for the 2D online bin packing problem, to maximize packing density. This method takes an image of the current bin state as input and determines the precise location for the next object placement. The reward function encourages placing objects in a way that maximizes space for future placements. The method is extendable to 3D online bin-packing problems.

For grouped 2D bin packing, common in industries like furniture manufacturing and glass cutting, where orders are divided into groups and optimized within each group, \cite{ao2023learning} presents a hierarchical reinforcement learning approach. The method was successfully developed in a Chinese factory, reducing the raw material costs. \cite{li2022one} proposes SAC with a recurrent attention encoder to capture inter-box dependencies and a conditional query decoder for reasoning about subsequent actions in 2D and 3D packing problems. This approach demonstrates superior space utilization compared to baselines, especially in offline and online strip packing scenarios.

To address uncertainties in real-world packing problems, \cite{zhang2022deep} presents a hybrid heuristic algorithm that combines enhanced scoring rules with a DQN, which dynamically selects heuristics through a data-driven process, to solve the truck routing and online 2D strip packing problem.

We can mention other works which combine RL with scoring rules. \cite{zhao2022reinforcement}, for instance, employed Q-learning for sequencing and the bottom-left centroid rule for positioning. Fang et al. \cite{fang2023deep} leveraged REINFORCE with the MaxRect-BL algorithm to exploit underlying packing patterns. It \cite{zhu2020hybrid} Reinforcement Learning-based Simple Random Algorithm (RSRA) algorithm, integrating skyline-based scoring rules with a DQN, has demonstrated effectiveness.

This section shows a range of RL methods applied to 2D regular packing problems. As research in this area advances, there is also an increasing focus on expanding 3D solutions~\cite{wu2021research,zhao2022learning,puche2022online,zuo2022three} and tackling irregular shapes~\cite{crescitelli2023deep,fang2023hybrid,fang2022research,fang2021reinforcement,yang2023learning}.

\section{DRL approach for 2D+1 packing problem}

This section describes our DRL solution for a 2D+1 packing environment, inspired by real-world scenarios related to aerospace composite manufacturing. The environment simulates the task of efficiently packing rectangular pieces onto two distinct boards with limited height. It was built using the OpenAI Gymnasium framework and represents the packing scenario with the following key components:

\begin{enumerate}
    \item \textbf{Observation space:}  It consists of two matrices, each one representing a board of length $X$ width dimensions. Additionally, four integer values are included, corresponding to the quantities of four different types of piece.
    
    \item \textbf{Action space:} It comprises a multi-discrete space, encompassing the $(x,y)$ coordinates for the top-left corner of a piece placement, an index selecting the target board, and another index specifying the piece to be chosen from the available set.
    
    \item \textbf{Algorithm} It is structured to reward the agent for positive actions that effectively fill the available spaces in the environment. Conversely, penalties are applied for invalid actions, such as selecting a piece with zero remaining quantity, attempting to place a piece on an already occupied coordinate, or putting a piece that exceeds the tooling cart's height. The process proceeds in Algorithm \ref{alg:stepfunction}.

   The agent and our simulator interact during each episode in a discrete-time sequence, $t = 0, 1, 2, 3, \cdots$. At each time step $t$, the agent is provided with a representation of the boards and the quantities of pieces to be placed, $s_{t} \in S$, where $S$ represents the set of available positions on the board and the piece's type. The action taken by the agent, denoted as $a_{t} \in A(s_{t})$, consists of selecting the coordinates $(x,y)$, the index board, and the index piece in-state $s_{t}$ for placement. At time step $t + 1$, as a result of this action, the agent receives a numerical reward $r_{t+1} \in R$ and transitions to a new state $s_{t+1}$, as shown in Figure~\ref{fig:rlmodel}.

    \begin{figure}[htbp!]
    \centerline{\includegraphics[width=6.4cm]{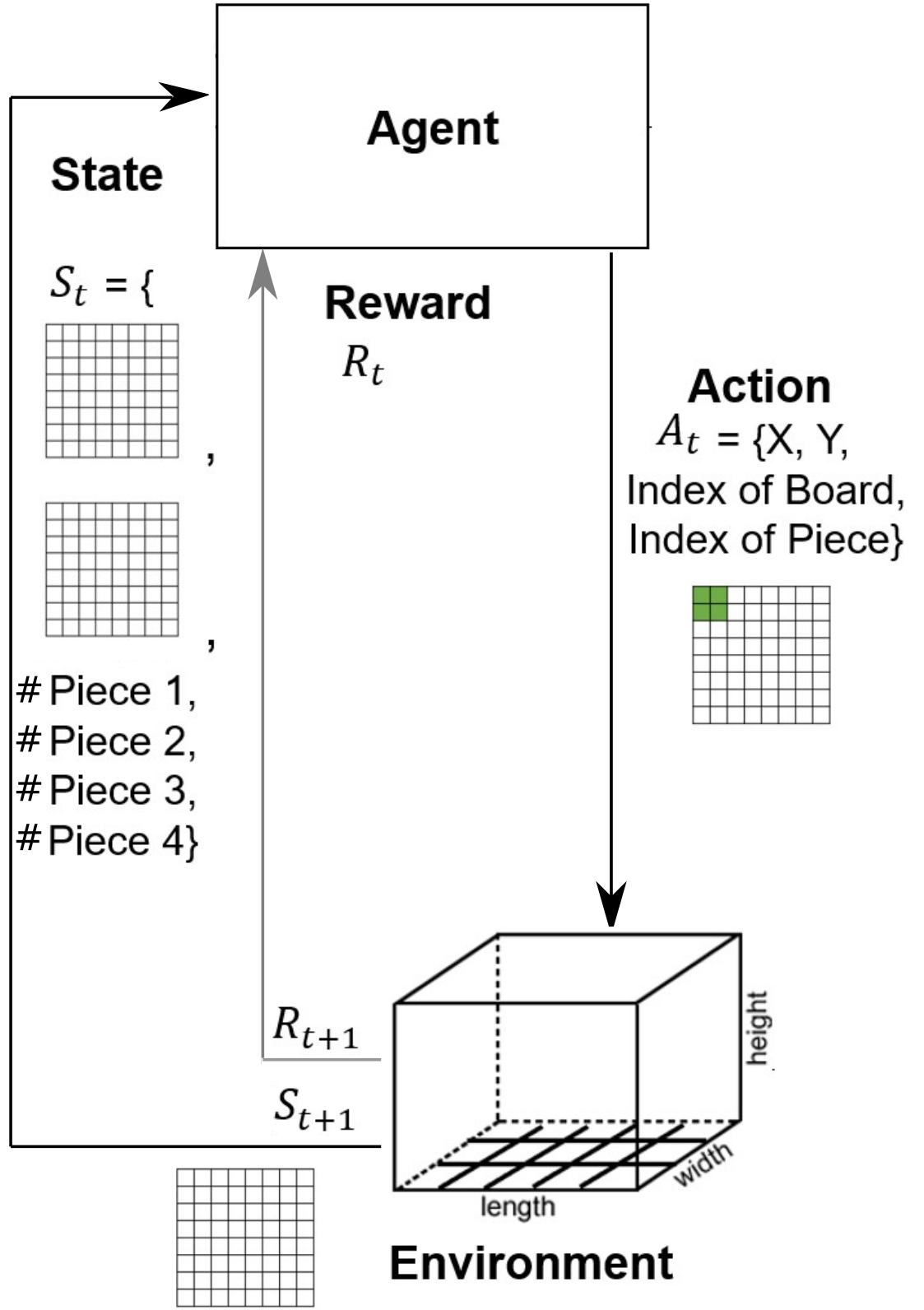}}
    \caption{Our simulator pipeline based on reinforcement learning methods.}
    \label{fig:rlmodel}
    \end{figure}

    The $R_{\text{height}}$ is determined by the following conditions:
    \begin{itemize}
        \item If $\frac{\text{piece\_height}}{\text{board\_height}} \times100\leq50$, then $R_{height}=0$
        \item Else if $\frac{\text{piece\_height}}{\text{board\_height}} \times100\leq80$, then $R_{height}=1$
        \item Else if $\frac{\text{piece\_height}}{\text{board\_height}} \times100\leq100$, then $R_{\text{height}}=2$ (Optimal)
        \item Else if $\frac{\text{piece\_height}}{\text{board\_height}} \times100>100$, then $R_{\text{height}}=-2$

    \end{itemize}

    \item \textbf{Training and Testing:} We only employed PPO and A2C methods in the Stable Baselines library, because they support multi-discrete action spaces. PPO, a state-of-the-art model-free reinforcement learning algorithm \cite{sun2019model}, is particularly effective in this context. Standard implementations of Deep Q-Network (DQN) and Soft Actor-Critic (SAC) are not directly applicable to multi-discrete action spaces.

    Our agents were trained in a 2D+1 packing environment for $10$ million episodes. Evaluations were conducted every $50$ episodes under deterministic conditions, and each experiment was repeated $10$ times. For both PPO and A2C, we used a linear learning rate of $0.0005$ and a discount factor (gamma) of $0.95$ as hyperparameters.

\end{enumerate}

\section{Experiments}
This section presents the experimental protocol and results achieved by DRL methods.

\subsection{Experimental Methodology}
In this work, we conducted six different experiments: three of them using even board dimensions ($8 \times 8$) and three using odd board dimensions ($7 \times 7$). Each set of experiments followed these conditions: (1) the pieces and boards were constrained to a uniform height, (2) the board $1$ was taller than the board $2$, and (3) the board $2$ was taller than the board $1$. Table~\ref{tab:setup_exp} summarizes the setup adopted in the experiments.

\begin{table}[htbp]
\centering
\caption{Setup of the experiments.}
\begin{tabular}{|c|c|c|c|c|c|}
\hline
Exp. ID & Even/Odd & Weight & Length & Height 0 & Height 1 \\ \hline
1             & Even     & 8      & 8      & 100               & 100               \\ \hline
2             & Even     & 8      & 8      & 120               & 80                \\ \hline
3             & Even     & 8      & 8      & 80                & 120               \\ \hline
4             & Odd      & 7      & 7      & 100               & 100               \\ \hline
5             & Odd      & 7      & 7      & 120               & 80                \\ \hline
6             & Odd      & 7      & 7      & 80                & 120               \\ \hline
\end{tabular}
\label{tab:setup_exp}
\end{table}

We also employed four types of pieces: $2 \times 2$ and $2 \times 1$, with heights of either 115 or 75 centimeters. These pieces are placed within two boards that define the environment's boundaries. Figure \ref{fig:pieces} shows their shapes.

\begin{figure}[htbp]
\centerline{\includegraphics[width=6cm]{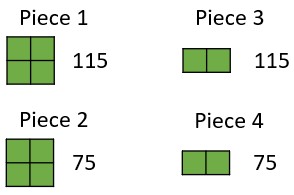}}
\caption{The types of piece available to be placed into the boards.}
\label{fig:pieces}
\end{figure}

Table~\ref{tab:pieces} shows the amount of pieces was used for each experiment.

\begin{table}[htbp]
\centering
\caption{The sets of pieces used for each experiment.}
\begin{tabular}{|c|c|c|c|c|}
\hline
Exp. ID & Qty Piece 1 & Qty Piece 2 & Qty Piece 3 & Qty Piece 4 \\ \hline
1 & 8 & 8 & 16 & 16 \\ \hline
2 & 8 & 8 & 16 & 16 \\ \hline
3 & 8 & 8 & 16 & 16 \\ \hline
4 & 6 & 6 & 9 & 9 \\ \hline
5 & 6 & 6 & 9 & 9 \\ \hline
6 & 6 & 6 & 9 & 9 \\ \hline
\end{tabular}
\label{tab:pieces}
\end{table}

In a real-world aerospace manufacturing setting, the number of parts in an autoclave can vary significantly based on available volume and batch size. We can expect around 30 to 50 parts per curing cycle. However, the number may be lower, such as when dealing with aircraft fairings. It is important to note that the packing phase does not involve irregular parts due to the safety margins necessary to achieve the desired product properties. Furthermore, since it operates on batches of parts, we should abstract these constraints, focusing on the packing process.

In the experiments carried out, the PPO, A2C and MaxRect-BL methods have been compared. We selected the MaxRect-BL approach, aligning with the bottom-left placement strategies employed in prior work by \cite{zhao2022reinforcement} and \cite{fang2023deep} within the context of RL}. To address the limitations of MaxRect-BL in handling height constraints, we implemented a modified version inspired by the BFDH. This modified approach prioritizes the height orientation of the pieces before considering their size during the packing process, effectively improving the packing efficiency. Furthermore, the simulations were performed on a Core i7 processor with 16 GB of RAM. Each PPO training session lasted approximately 7 hours.

\subsection{Even Experiments}
\begin{figure}[htbp!]
\centerline{\includegraphics[width=10cm]{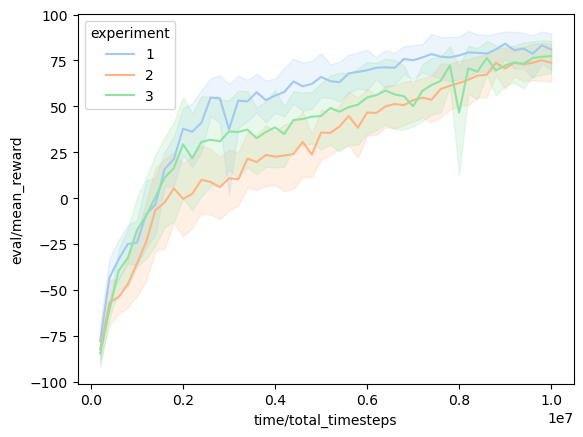}}
\caption{Evaluation curves of the three even experiments using PPO.}
\label{fig:evaluate_curve_even}
\end{figure}

Figure~\ref{fig:evaluate_curve_even} illustrates the evaluation curves for 10 independent PPO runs across the three experimental conditions. These experiments demonstrated optimal performance by achieving the maximum reward through $100\%$ correct board fillings. Negative reward values signify incorrect board configurations.

\begin{figure}[htbp!]
\centerline{\includegraphics[width=10cm]{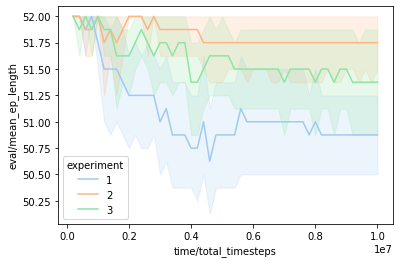}}
\caption{Mean episode length over even experiments using PPO.}
\label{fig:lenght_curve_even}
\end{figure}

{Figure~\ref{fig:lenght_curve_even} presents the curves depicting the mean episode length across $10$ independent PPO runs under the three experimental conditions. These results highlight the progression of the mean episode length throughout iterations, providing insights into the agent's performance dynamics and its efficiency in solving the problem. The optimal episode length occurs approximately when the maximum number of pieces is successfully packed onto a board.}

In the experiments carried out, the A2C method proved to be less practical than the PPO due to its significantly longer convergence time and greater instability during training. Furthermore, A2C often fails to achieve optimal performance compared to PPO method. We selected 4 results from each experimental setup to compare these RL methods. Table~\ref{tab:rl_1} contains the mean percentage of correct board fills (Mean) and its standard deviation (Std).

\begin{table}[ht!]
\centering
\caption{Comparative analysis between PPO and A2C methods for correctly board filling.}
\begin{tabular}{|c|cc|cc|}
\hline
\multirow{2}{*}{Experiment} & \multicolumn{2}{c|}{PPO}         & \multicolumn{2}{c|}{A2C}         \\ \cline{2-5} 
                            & \multicolumn{1}{c|}{Mean } & Std  & \multicolumn{1}{c|}{Mean } & std  \\ \hline
1                           & \multicolumn{1}{c|}{96.0\%} & 3.0\% & \multicolumn{1}{c|}{88.0\%} & 6.0\% \\ \hline
2                           & \multicolumn{1}{c|}{96.0\%} & 5.0\% & \multicolumn{1}{c|}{34.0\%} & 23.0\% \\ \hline
3                           & \multicolumn{1}{c|}{94.0\%} & 5.0\% & \multicolumn{1}{c|}{74.0\%} & 5.0\% \\ \hline
\end{tabular}
\label{tab:rl_1}
\end{table}

\subsubsection{Experiment 1}
\begin{figure}[htbp]
\centerline{\includegraphics[width=5cm]{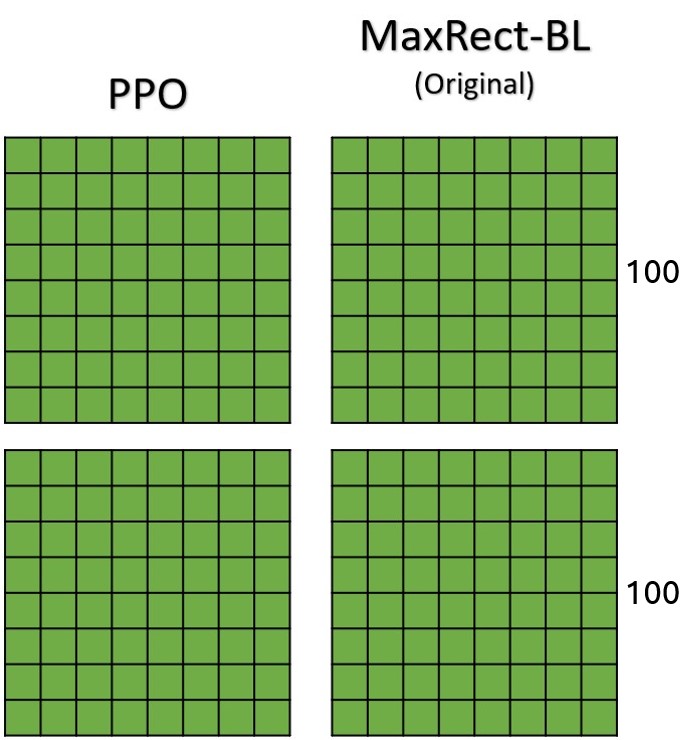}}
\caption{Experiment 1 - All of pieces and boards were constrained to a uniform height for this experiment.}
\label{fig:experiment1}
\end{figure}

All of pieces and boards were constrained to a uniform height for this experiment. Both the PPO and MaxRect-Bl algorithms achieved complete coverage (100\%) of the boards, as demonstrated in Figure \ref{fig:experiment1}. The green regions highlight the optimal placements determined by the algorithms during the packing process.

\subsubsection{Experiment 2}
\begin{figure}[htbp]
\centerline{\includegraphics[width=5cm]{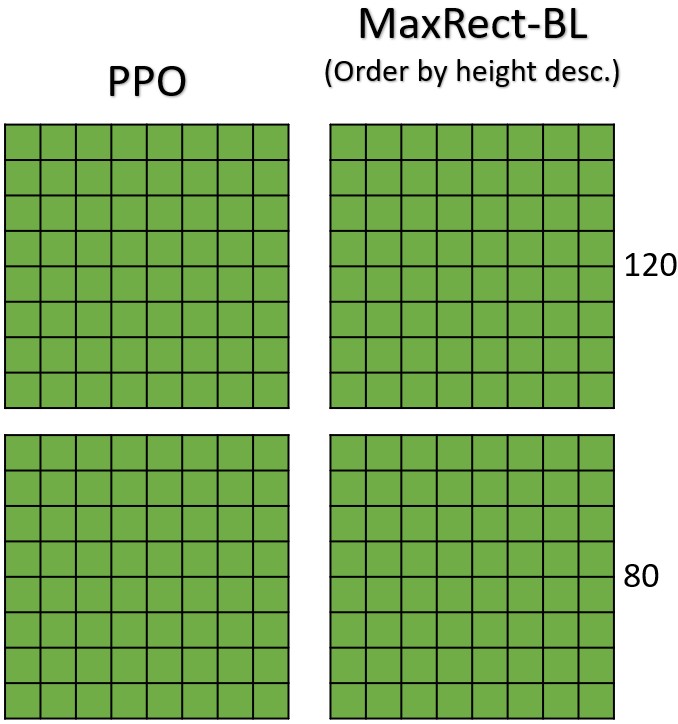}}
\caption{Experiment 2 - board 1 taller than board 2.}
\label{fig:experiment2}
\end{figure}

Both PPO and MaxRect-BL achieved 100\% coverage. However, MaxRect-BL's optimal performance was contingent on a specific piece sorting strategy: first by descending height, then by descending dimensions. The BFDH heuristic could also achieve optimal performance. Figure~\ref{fig:experiment2} shows the convergence behavior of the experiment.

\begin{figure}[htbp]
\centerline{\includegraphics[width=5cm]{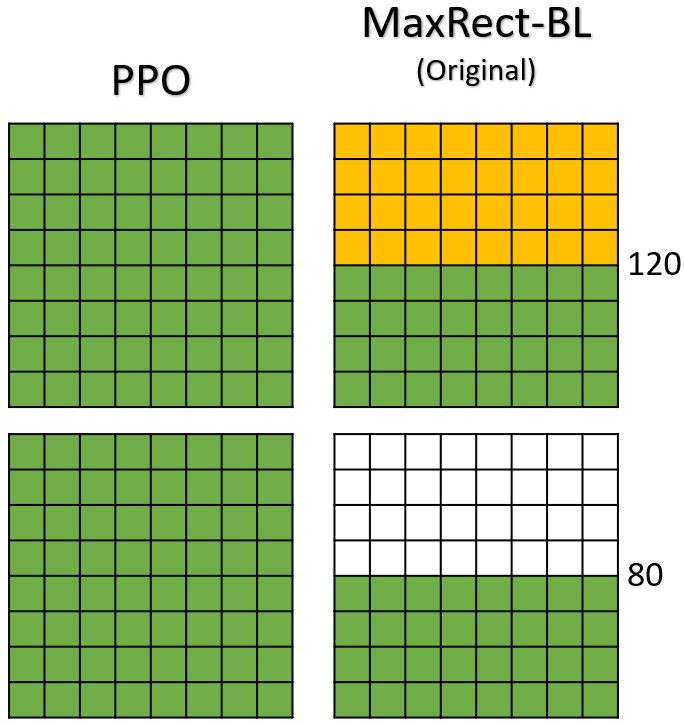}}
\caption{Experiment 2 without order pieces for MaxRect-BL.}
\label{fig:experiment2b}
\end{figure}

Without sorting by height, the MaxRect-BL fails to converge as effectively, as indicated in Figure~\ref{fig:experiment2b}. This occurs because MaxRect-BL initially places 8 Pieces\_1 (with $R_{height} = 2$) and 8 Pieces\_2 (with $R_{height} = 1$), which fill all the available space on board 0. It then attempts to add 16 Pieces\_3 (whose height exceeds the board's height, causing them to be skipped) and finally places 16 Pieces\_4 (with $R_{height} = 2$) on board 1. The yellow areas highlight suboptimal placement choices resulting from $R_{height} < 2$, while the white areas represent unused space on the board.

\subsubsection{Experiment 3} 
\begin{figure}[htbp]
\centerline{\includegraphics[width=5cm]{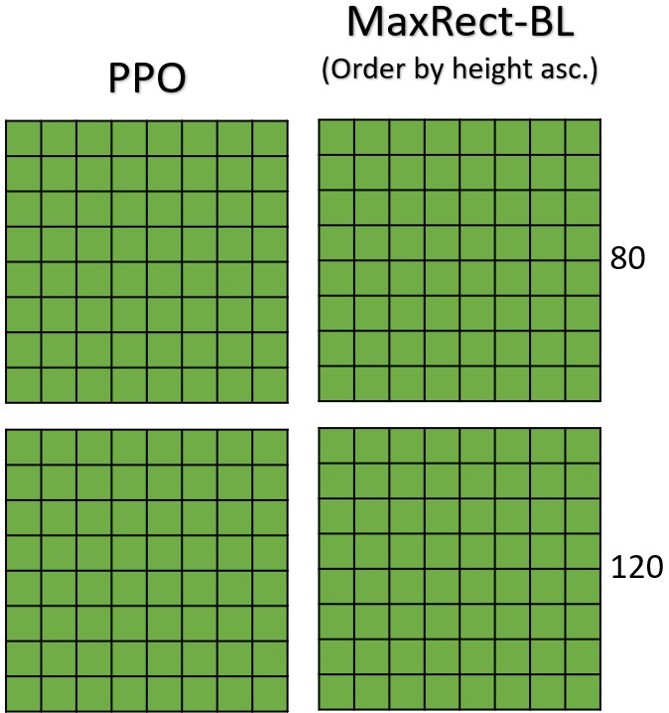}}
\caption{Experiment 3 - board 2 taller than board 1.}
\label{fig:experiment3}
\end{figure}

In this experiment, both PPO and MaxRect-BL achieved 100\% coverage. While MaxRect-BL required a specific piece sorting strategy (ascending height, descending dimensions) for optimal performance, the BFDH heuristic could also achieve optimal results in this scenario. Figure~\ref{fig:experiment3} illustrates the convergence behavior of the experiment.

\begin{figure}[htbp]
\centerline{\includegraphics[width=5cm]{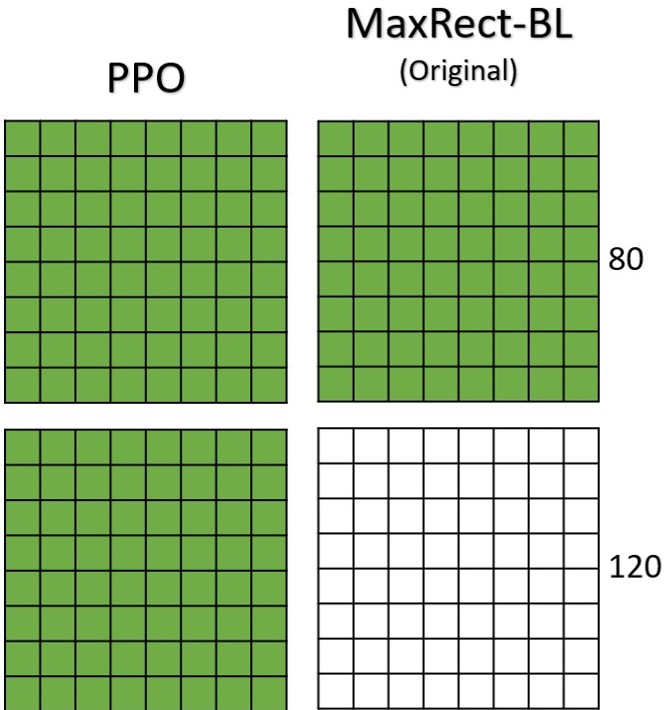}}
\caption{Experiment 3 without order pieces for MaxRect-BL}
\label{fig:experiment3b}
\end{figure}

Without sorting by height, the MaxRect-BL fails to converge as effectively, as indicated in Figure~\ref{fig:experiment3b}. This occurs because MaxRect-BL first attempts to place 8 Pieces\_1 (whose height exceeds the board's limit, causing them to be skipped), but successfully adds 8 Pieces\_2 (with $R_{height} = 2$). It then tries to add 16 Pieces\_3 (again skipped due to their excessive height) and finally places 16 Pieces\_4 (which meet the optimal condition of $R_{\text{height}} = 2$) on board 0.

\subsection{Odd boards}
\begin{figure}[htbp]
\centerline{\includegraphics[width=10cm]{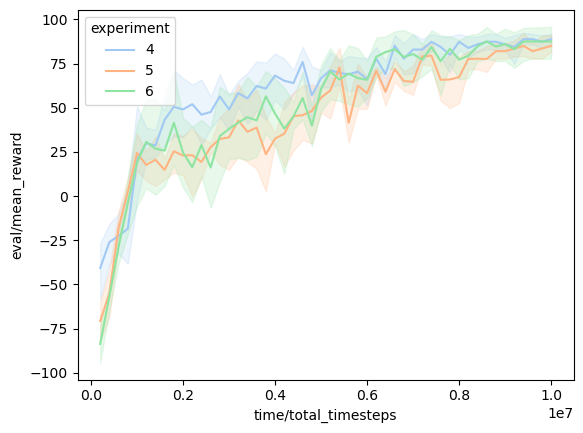}}
\caption{Evaluation curves of the three odd experiments using PPO.}
\label{fig:evaluate_curve_odd}
\end{figure}

Figure~\ref{fig:evaluate_curve_odd} illustrates the evaluation curves for 10 independent PPO runs across the three experimental conditions. These experiments demonstrated optimal performance by achieving the maximum reward through 100\% correct board fillings. Negative reward values signify incorrect board configurations.

\begin{figure}[htbp]
\centerline{\includegraphics[width=10cm]{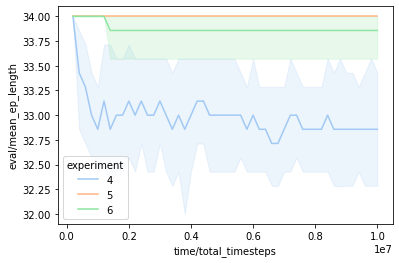}}
\caption{Mean episode length over odd experiments using
PPO.}
\label{fig:lenght_curve_odd}
\end{figure}

{Figure~\ref{fig:lenght_curve_odd} presents the curves depicting the mean episode length across $10$ independent PPO runs under the three experimental conditions. These results highlight the progression of the mean episode length throughout iterations, providing insights into the agent's performance dynamics and its efficiency in solving the problem. The optimal episode length occurs approximately when the maximum number of pieces is successfully packed onto a board.}

In these experiments, it was again possible to observe that A2C method achieved an inferior performance when compared to PPO method. We selected 4 results from each experimental condition to compare these RL methods. Table \ref{rl_2} contains the mean percentage of correct board fills (Mean) and its standard deviation (Std).

\begin{table}[ht!]
\centering
\caption{Comparative analysis between PPO and A2C methods for correct board fills.}

\begin{tabular}{|c|cc|cc|}
\hline
\multirow{2}{*}{experiment} & \multicolumn{2}{c|}{PPO}         & \multicolumn{2}{c|}{A2C}         \\ \cline{2-5} 
                            & \multicolumn{1}{c|}{Mean} & Std  & \multicolumn{1}{c|}{Mean} & Std  \\ \hline
4                           & \multicolumn{1}{c|}{97.0\%} & 3.0\% & \multicolumn{1}{c|}{82.0\%}   & 8.0\% \\ \hline
5                           & \multicolumn{1}{c|}{97.0\%} & 3.0\% & \multicolumn{1}{c|}{88.0\%}   & 4.0\% \\ \hline
6                           & \multicolumn{1}{c|}{97.0\%} & 4.0\% & \multicolumn{1}{c|}{81.0\%}   & 8.0\% \\ \hline
\end{tabular}
\label{rl_2}
\end{table}

\subsubsection{Experiment 4}
\begin{figure}[htbp]
\centerline{\includegraphics[width=5cm]{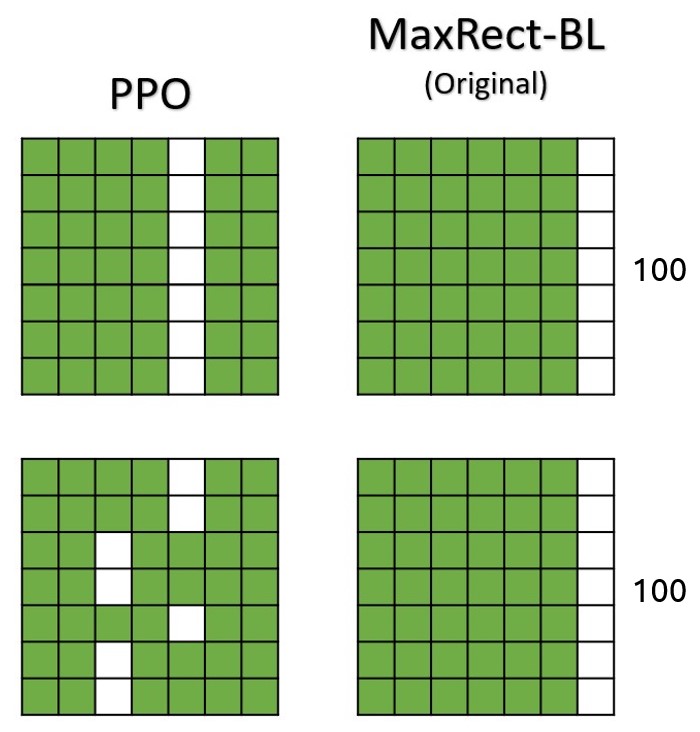}}
\caption{Experiment 4 - All of pieces and boards were constrained to a uniform height for this experiment.}
\label{fig:experiment4}
\end{figure}

All of pieces and boards were constrained to a uniform height for this experiment. Both the PPO and MaxRect-Bl algorithms achieved complete coverage (100\%) of the boards, as demonstrated in Figure \ref{fig:experiment4}. The green regions highlight the optimal placements determined by the algorithms during the packing process.

\subsubsection{Experiment 5}
\begin{figure}[htbp]
\centerline{\includegraphics[width=5cm]{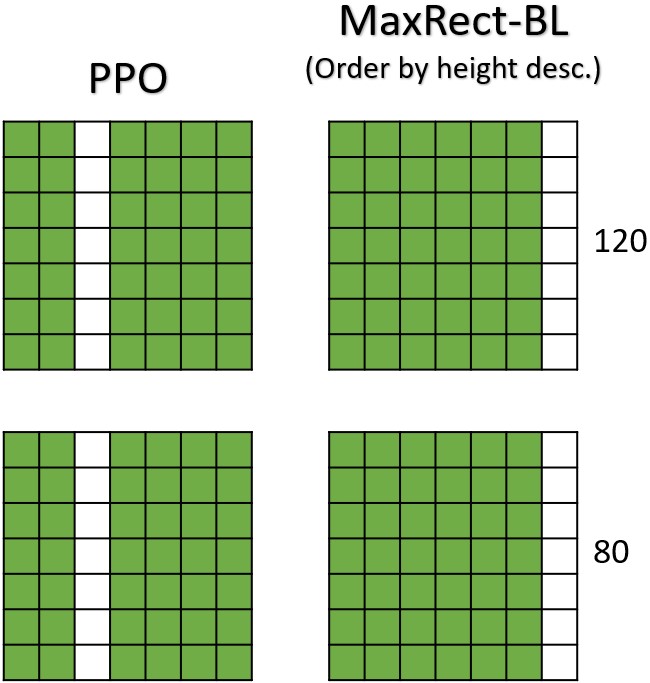}}
\caption{Experiment 5 - board 1 taller than board 2.}
\label{fig:experiment5}
\end{figure}

PPO and MaxRect-Bl (ordered by height descending) successfully placed all pieces. MaxRect-BL exhibits the same limitations as in Experiment 2 without this ordering. Figure~\ref{fig:experiment5} shows the convergence behavior of the experiment.

\subsubsection{Experiment 6}
\begin{figure}[htbp]
\centerline{\includegraphics[width=5cm]{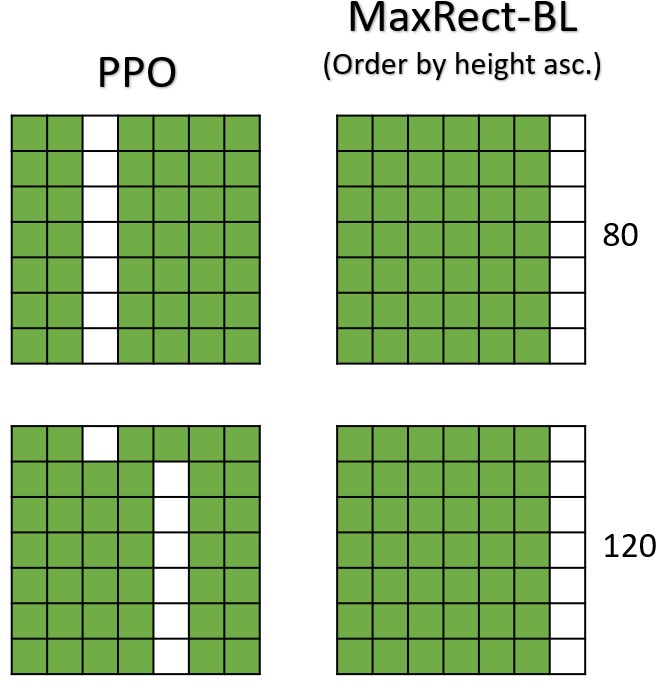}}
\caption{Experiment 6 - board 2 taller than board 1.}
\label{fig:experiment6}
\end{figure}

PPO and MaxRect-Bl (ordered by height ascending) successfully placed all pieces. MaxRect-BL exhibits the same limitations as in Experiment 3 without this ordering. Figure~\ref{fig:experiment6} shows the convergence behavior of the experiment.

\section{Conclusions}
This paper proposed a 2D+1 simulator, and developed a spatially constrained packing problem within the OpenAI Gymnasium framework, for the packing problem in the offline approach. This simulator employs an observation space comprising two boards and four different types of pieces and their associated quantities. It supports multi-discrete action space, allowing the selection of a position on a specific board and the choice of a piece to place. Furthermore, this paper introduced a new spatial-variant reward function that maximizes coverage by considering both dimension and height of the pieces. This research conducted a literature review focused on deep reinforcement learning solutions for the 2D regular packing problem. Since 2018, publications involving DRL for this type of problem have attracted the attention of researchers; however, there are still research gaps, such as the use of on-policy actor-critic methods for the target task. In the performed experiments, it was possible to observe that PPO and MaxRect-BL (with height ordering) have correctly allocated all of the pieces. However, MaxRect-BL without height ordering exhibited poorer performance, as illustrated in Figures \ref{fig:experiment2b} and \ref{fig:experiment3b}. As the problem complexity increases (e.g., multiple boards), the effectiveness of simple heuristics like height-based ordering diminishes. While the BFDH heuristic is viable for packing items, PPO's ability to learn and adapt dynamically through exploration and exploitation provides a more flexible and potentially superior solution. The A2C did not show better results than PPO in the experiments. As future work, to enhance the simulator's fidelity as a digital representation of an aerospace industry autoclave for composite material curing, we plan to implement key improvements, including material allocation constraints to ensure accurate material placement based on specific curing types, thus reflecting real-world production processes. Additionally, we will integrate thermocouple and pressure sensor simulations to capture precise temperature and pressure conditions within the autoclave, providing valuable data for process optimization and quality control. Furthermore, a mechanism will be added to simulate material delivery deadlines, ensuring the simulator reflects the time-sensitive nature of production operations. These enhancements will result in a more comprehensive and realistic model of the autoclave curing process, enabling engineers to conduct more effective simulations and optimize production workflows.

\section*{Acknowledgements}
The authors would like to thank the National Council for Scientific and Technological Development (CNPq) for granting a scholarship to Victor Pugliese through the Academic Master's and Doctorate Program in Innovation (MAI/DAI) in collaboration with the EMBRAER S.A. company.

\bibliographystyle{elsarticle-harv}
\bibliography{bib}

\begin{thebibliography}{37}
\expandafter\ifx\csname natexlab\endcsname\relax\def\natexlab#1{#1}\fi
\providecommand{\url}[1]{\texttt{#1}}
\providecommand{\href}[2]{#2}
\providecommand{\path}[1]{#1}
\providecommand{\DOIprefix}{doi:}
\providecommand{\ArXivprefix}{arXiv:}
\providecommand{\URLprefix}{URL: }
\providecommand{\Pubmedprefix}{pmid:}
\providecommand{\doi}[1]{\href{http://dx.doi.org/#1}{\path{#1}}}
\providecommand{\Pubmed}[1]{\href{pmid:#1}{\path{#1}}}
\providecommand{\bibinfo}[2]{#2}
\ifx\xfnm\relax \def\xfnm[#1]{\unskip,\space#1}\fi
\bibitem[{Alem{\~a}o et~al.(2021)Alem{\~a}o, Rocha and
  Barata}]{alemao2021smart}
\bibinfo{author}{Alem{\~a}o, D.}, \bibinfo{author}{Rocha, A.D.},
  \bibinfo{author}{Barata, J.}, \bibinfo{year}{2021}.
\newblock \bibinfo{title}{Smart manufacturing scheduling
  approaches—systematic review and future directions}.
\newblock \bibinfo{journal}{Applied Sciences} \bibinfo{volume}{11},
  \bibinfo{pages}{2186}.
\bibitem[{Ao et~al.(2023)Ao, Zhang, Li and Jin}]{ao2023learning}
\bibinfo{author}{Ao, W.}, \bibinfo{author}{Zhang, G.}, \bibinfo{author}{Li,
  Y.}, \bibinfo{author}{Jin, D.}, \bibinfo{year}{2023}.
\newblock \bibinfo{title}{Learning to solve grouped 2d bin packing problems in
  the manufacturing industry}, in: \bibinfo{booktitle}{Proceedings of the 29th
  ACM SIGKDD Conference on Knowledge Discovery and Data Mining}, pp.
  \bibinfo{pages}{3713--3723}.
\bibitem[{Azami(2016)}]{azami2016scheduling}
\bibinfo{author}{Azami, A.}, \bibinfo{year}{2016}.
\newblock \bibinfo{title}{Scheduling Hybrid Flow Lines of Aerospace Composite
  Manufacturing Systems}.
\newblock Ph.D. thesis. Concordia University.
\bibitem[{Azami et~al.(2018)Azami, Demirli and Bhuiyan}]{azami2018scheduling}
\bibinfo{author}{Azami, A.}, \bibinfo{author}{Demirli, K.},
  \bibinfo{author}{Bhuiyan, N.}, \bibinfo{year}{2018}.
\newblock \bibinfo{title}{Scheduling in aerospace composite manufacturing
  systems: a two-stage hybrid flow shop problem}.
\newblock \bibinfo{journal}{The International Journal of Advanced Manufacturing
  Technology} \bibinfo{volume}{95}, \bibinfo{pages}{3259--3274}.
\bibitem[{Cheng et~al.(2021)Cheng, Kolobov and
  Swaminathan}]{cheng2021heuristic}
\bibinfo{author}{Cheng, C.A.}, \bibinfo{author}{Kolobov, A.},
  \bibinfo{author}{Swaminathan, A.}, \bibinfo{year}{2021}.
\newblock \bibinfo{title}{Heuristic-guided reinforcement learning}.
\newblock \bibinfo{journal}{Advances in Neural Information Processing Systems}
  \bibinfo{volume}{34}, \bibinfo{pages}{13550--13563}.
\bibitem[{Collart(2015)}]{collart2015application}
\bibinfo{author}{Collart, A.}, \bibinfo{year}{2015}.
\newblock \bibinfo{title}{An application of mathematical optimization to
  autoclave packing and scheduling in a composites manufacturing facility} .
\bibitem[{Crescitelli and Oshima(2023)}]{crescitelli2023deep}
\bibinfo{author}{Crescitelli, V.}, \bibinfo{author}{Oshima, T.},
  \bibinfo{year}{2023}.
\newblock \bibinfo{title}{A deep reinforcement learning method for 2d irregular
  packing with dense reward}, in: \bibinfo{booktitle}{2023 Fifth International
  Conference on Transdisciplinary AI (TransAI)}, \bibinfo{organization}{IEEE}.
  pp. \bibinfo{pages}{270--271}.
\bibitem[{Elkington et~al.(2015)Elkington, Bloom, Ward, Chatzimichali and
  Potter}]{elkington2015hand}
\bibinfo{author}{Elkington, M.}, \bibinfo{author}{Bloom, D.},
  \bibinfo{author}{Ward, C.}, \bibinfo{author}{Chatzimichali, A.},
  \bibinfo{author}{Potter, K.}, \bibinfo{year}{2015}.
\newblock \bibinfo{title}{Hand layup: understanding the manual process}.
\newblock \bibinfo{journal}{Advanced manufacturing: polymer \& composites
  science} \bibinfo{volume}{1}, \bibinfo{pages}{138--151}.
\bibitem[{Fang et~al.(2022)Fang, Rao, Ding and Meng}]{fang2022research}
\bibinfo{author}{Fang, J.}, \bibinfo{author}{Rao, Y.}, \bibinfo{author}{Ding,
  W.}, \bibinfo{author}{Meng, R.}, \bibinfo{year}{2022}.
\newblock \bibinfo{title}{Research on two-dimensional intelligent nesting based
  on sarsa-learning}, in: \bibinfo{booktitle}{2022 5th International Conference
  on Advanced Electronic Materials, Computers and Software Engineering
  (AEMCSE)}, \bibinfo{organization}{IEEE}. pp. \bibinfo{pages}{826--829}.
\bibitem[{Fang et~al.(2021)Fang, Rao, Guo and Zhao}]{fang2021reinforcement}
\bibinfo{author}{Fang, J.}, \bibinfo{author}{Rao, Y.}, \bibinfo{author}{Guo,
  X.}, \bibinfo{author}{Zhao, X.}, \bibinfo{year}{2021}.
\newblock \bibinfo{title}{A reinforcement learning algorithm for
  two-dimensional irregular packing problems}, in:
  \bibinfo{booktitle}{Proceedings of the 2021 4th International Conference on
  Algorithms, Computing and Artificial Intelligence}, pp.
  \bibinfo{pages}{1--6}.
\bibitem[{Fang et~al.(2023a)Fang, Rao and Shi}]{fang2023deep}
\bibinfo{author}{Fang, J.}, \bibinfo{author}{Rao, Y.}, \bibinfo{author}{Shi,
  M.}, \bibinfo{year}{2023}a.
\newblock \bibinfo{title}{A deep reinforcement learning algorithm for the
  rectangular strip packing problem}.
\newblock \bibinfo{journal}{Plos one} \bibinfo{volume}{18},
  \bibinfo{pages}{e0282598}.
\bibitem[{Fang et~al.(2023b)Fang, Rao, Zhao and Du}]{fang2023hybrid}
\bibinfo{author}{Fang, J.}, \bibinfo{author}{Rao, Y.}, \bibinfo{author}{Zhao,
  X.}, \bibinfo{author}{Du, B.}, \bibinfo{year}{2023}b.
\newblock \bibinfo{title}{A hybrid reinforcement learning algorithm for 2d
  irregular packing problems}.
\newblock \bibinfo{journal}{Mathematics} \bibinfo{volume}{11},
  \bibinfo{pages}{327}.
\bibitem[{Haskilic et~al.(2023)Haskilic, Ulucan, Atici and
  Sarac}]{haskilic2023real}
\bibinfo{author}{Haskilic, V.}, \bibinfo{author}{Ulucan, A.},
  \bibinfo{author}{Atici, K.B.}, \bibinfo{author}{Sarac, S.B.},
  \bibinfo{year}{2023}.
\newblock \bibinfo{title}{A real-world case of autoclave loading and scheduling
  problems in aerospace composite material production}.
\newblock \bibinfo{journal}{Omega} , \bibinfo{pages}{102918}.
\bibitem[{Huang et~al.(2022)Huang, Kanervisto, Raffin, Wang, Onta{\~n}{\'o}n
  and Dossa}]{huang2022a2c}
\bibinfo{author}{Huang, S.}, \bibinfo{author}{Kanervisto, A.},
  \bibinfo{author}{Raffin, A.}, \bibinfo{author}{Wang, W.},
  \bibinfo{author}{Onta{\~n}{\'o}n, S.}, \bibinfo{author}{Dossa, R.F.J.},
  \bibinfo{year}{2022}.
\newblock \bibinfo{title}{A2c is a special case of ppo}.
\newblock \bibinfo{journal}{arXiv preprint arXiv:2205.09123} .
\bibitem[{Kaelbling et~al.(1996)Kaelbling, Littman and
  Moore}]{kaelbling1996reinforcement}
\bibinfo{author}{Kaelbling, L.P.}, \bibinfo{author}{Littman, M.L.},
  \bibinfo{author}{Moore, A.W.}, \bibinfo{year}{1996}.
\newblock \bibinfo{title}{Reinforcement learning: A survey}.
\newblock \bibinfo{journal}{Journal of artificial intelligence research}
  \bibinfo{volume}{4}, \bibinfo{pages}{237--285}.
\bibitem[{Keras(2022)}]{ppokeras}
\bibinfo{author}{Keras, F.}, \bibinfo{year}{2022}.
\newblock \bibinfo{title}{{PPO} proximal policy optimization}.
\newblock \URLprefix \url{https://keras.io/examples/rl/ppo_cartpole/}.
\bibitem[{Kundu et~al.(2019)Kundu, Dutta and Kumar}]{kundu2019deep}
\bibinfo{author}{Kundu, O.}, \bibinfo{author}{Dutta, S.},
  \bibinfo{author}{Kumar, S.}, \bibinfo{year}{2019}.
\newblock \bibinfo{title}{Deep-pack: A vision-based 2d online bin packing
  algorithm with deep reinforcement learning}, in: \bibinfo{booktitle}{2019
  28th IEEE International Conference on Robot and Human Interactive
  Communication (RO-MAN)}, \bibinfo{organization}{IEEE}. pp.
  \bibinfo{pages}{1--7}.
\bibitem[{Li et~al.(2022)Li, Gu, Wang, Ren and Lau}]{li2022one}
\bibinfo{author}{Li, D.}, \bibinfo{author}{Gu, Z.}, \bibinfo{author}{Wang, Y.},
  \bibinfo{author}{Ren, C.}, \bibinfo{author}{Lau, F.C.}, \bibinfo{year}{2022}.
\newblock \bibinfo{title}{One model packs thousands of items with recurrent
  conditional query learning}.
\newblock \bibinfo{journal}{Knowledge-Based Systems} \bibinfo{volume}{235},
  \bibinfo{pages}{107683}.
\bibitem[{Mnih et~al.(2016)Mnih, Badia, Mirza, Graves, Lillicrap, Harley,
  Silver and Kavukcuoglu}]{mnih2016asynchronous}
\bibinfo{author}{Mnih, V.}, \bibinfo{author}{Badia, A.P.},
  \bibinfo{author}{Mirza, M.}, \bibinfo{author}{Graves, A.},
  \bibinfo{author}{Lillicrap, T.}, \bibinfo{author}{Harley, T.},
  \bibinfo{author}{Silver, D.}, \bibinfo{author}{Kavukcuoglu, K.},
  \bibinfo{year}{2016}.
\newblock \bibinfo{title}{Asynchronous methods for deep reinforcement
  learning}, in: \bibinfo{booktitle}{International conference on machine
  learning}, \bibinfo{organization}{PMLR}. pp. \bibinfo{pages}{1928--1937}.
\bibitem[{Oliveira et~al.(2016)Oliveira, Neuenfeldt, Silva and
  Carravilla}]{oliveira2016survey}
\bibinfo{author}{Oliveira, J.F.}, \bibinfo{author}{Neuenfeldt, A.},
  \bibinfo{author}{Silva, E.}, \bibinfo{author}{Carravilla, M.A.},
  \bibinfo{year}{2016}.
\newblock \bibinfo{title}{A survey on heuristics for the two-dimensional
  rectangular strip packing problem}.
\newblock \bibinfo{journal}{Pesquisa Operacional} \bibinfo{volume}{36},
  \bibinfo{pages}{197--226}.
\bibitem[{Puche and Lee(2022)}]{puche2022online}
\bibinfo{author}{Puche, A.V.}, \bibinfo{author}{Lee, S.}, \bibinfo{year}{2022}.
\newblock \bibinfo{title}{Online 3d bin packing reinforcement learning solution
  with buffer}, in: \bibinfo{booktitle}{2022 ieee/rsj international conference
  on intelligent robots and systems (iros)}, \bibinfo{organization}{IEEE}. pp.
  \bibinfo{pages}{8902--8909}.
\bibitem[{Ramezankhani et~al.(2021)Ramezankhani, Crawford, Narayan,
  Voggenreiter, Seethaler and Milani}]{ramezankhani2021making}
\bibinfo{author}{Ramezankhani, M.}, \bibinfo{author}{Crawford, B.},
  \bibinfo{author}{Narayan, A.}, \bibinfo{author}{Voggenreiter, H.},
  \bibinfo{author}{Seethaler, R.}, \bibinfo{author}{Milani, A.S.},
  \bibinfo{year}{2021}.
\newblock \bibinfo{title}{Making costly manufacturing smart with transfer
  learning under limited data: A case study on composites autoclave
  processing}.
\newblock \bibinfo{journal}{Journal of Manufacturing Systems}
  \bibinfo{volume}{59}, \bibinfo{pages}{345--354}.
\bibitem[{S{\'a}enz~Imbacu{\'a}n(2020)}]{saenzevaluating}
\bibinfo{author}{S{\'a}enz~Imbacu{\'a}n, R.}, \bibinfo{year}{2020}.
\newblock \bibinfo{title}{Evaluating the impact of curriculum learning on the
  training process for an intelligent agent in a video game} .
\bibitem[{Schulman et~al.(2017)Schulman, Wolski, Dhariwal, Radford and
  Klimov}]{schulman2017proximal}
\bibinfo{author}{Schulman, J.}, \bibinfo{author}{Wolski, F.},
  \bibinfo{author}{Dhariwal, P.}, \bibinfo{author}{Radford, A.},
  \bibinfo{author}{Klimov, O.}, \bibinfo{year}{2017}.
\newblock \bibinfo{title}{Proximal policy optimization algorithms}.
\newblock \bibinfo{journal}{arXiv preprint arXiv:1707.06347} .
\bibitem[{Seizinger(2018)}]{seizinger2018two}
\bibinfo{author}{Seizinger, M.}, \bibinfo{year}{2018}.
\newblock \bibinfo{title}{The two dimensional bin packing problem with side
  constraints}, in: \bibinfo{booktitle}{Operations Research Proceedings 2017:
  Selected Papers of the Annual International Conference of the German
  Operations Research Society (GOR), Freie Universi{\"a}t Berlin, Germany,
  September 6-8, 2017}, pp. \bibinfo{pages}{45--50}.
\bibitem[{Sun et~al.(2019)Sun, Yuan, Liu and Sun}]{sun2019model}
\bibinfo{author}{Sun, Y.}, \bibinfo{author}{Yuan, X.}, \bibinfo{author}{Liu,
  W.}, \bibinfo{author}{Sun, C.}, \bibinfo{year}{2019}.
\newblock \bibinfo{title}{Model-based reinforcement learning via proximal
  policy optimization}, in: \bibinfo{booktitle}{2019 Chinese Automation
  Congress (CAC)}, \bibinfo{organization}{IEEE}. pp.
  \bibinfo{pages}{4736--4740}.
\bibitem[{Sutton and Barto(2018)}]{sutton2018reinforcement}
\bibinfo{author}{Sutton, R.S.}, \bibinfo{author}{Barto, A.G.},
  \bibinfo{year}{2018}.
\newblock \bibinfo{title}{Reinforcement learning: An introduction}.
\newblock \bibinfo{publisher}{MIT press}.
\bibitem[{Wang et~al.(2022)Wang, Thomas, Piechocki, Kapoor,
  Santos-Rodr{\'\i}guez and Parekh}]{wang2022self}
\bibinfo{author}{Wang, X.}, \bibinfo{author}{Thomas, J.D.},
  \bibinfo{author}{Piechocki, R.J.}, \bibinfo{author}{Kapoor, S.},
  \bibinfo{author}{Santos-Rodr{\'\i}guez, R.}, \bibinfo{author}{Parekh, A.},
  \bibinfo{year}{2022}.
\newblock \bibinfo{title}{Self-play learning strategies for resource assignment
  in open-ran networks}.
\newblock \bibinfo{journal}{Computer Networks} \bibinfo{volume}{206},
  \bibinfo{pages}{108682}.
\bibitem[{Wu and Yao(2021)}]{wu2021research}
\bibinfo{author}{Wu, Y.}, \bibinfo{author}{Yao, L.}, \bibinfo{year}{2021}.
\newblock \bibinfo{title}{Research on the problem of 3d bin packing under
  incomplete information based on deep reinforcement learning}, in:
  \bibinfo{booktitle}{2021 International conference on e-commerce and
  e-management (ICECEM)}, \bibinfo{organization}{IEEE}. pp.
  \bibinfo{pages}{38--42}.
\bibitem[{Xia et~al.(2021)Xia, Sacco, Kirkpatrick, Saidy, Nguyen, Kircaliali
  and Harik}]{xia2021digital}
\bibinfo{author}{Xia, K.}, \bibinfo{author}{Sacco, C.},
  \bibinfo{author}{Kirkpatrick, M.}, \bibinfo{author}{Saidy, C.},
  \bibinfo{author}{Nguyen, L.}, \bibinfo{author}{Kircaliali, A.},
  \bibinfo{author}{Harik, R.}, \bibinfo{year}{2021}.
\newblock \bibinfo{title}{A digital twin to train deep reinforcement learning
  agent for smart manufacturing plants: Environment, interfaces and
  intelligence}.
\newblock \bibinfo{journal}{Journal of Manufacturing Systems}
  \bibinfo{volume}{58}, \bibinfo{pages}{210--230}.
\bibitem[{Xie et~al.(2020)Xie, Zheng and Wu}]{xie2020two}
\bibinfo{author}{Xie, N.}, \bibinfo{author}{Zheng, S.}, \bibinfo{author}{Wu,
  Q.}, \bibinfo{year}{2020}.
\newblock \bibinfo{title}{Two-dimensional packing algorithm for autoclave
  molding scheduling of aeronautical composite materials production}.
\newblock \bibinfo{journal}{Computers \& Industrial Engineering}
  \bibinfo{volume}{146}, \bibinfo{pages}{106599}.
\bibitem[{Yang et~al.(2023)Yang, Pan, Li, Wu and Gao}]{yang2023learning}
\bibinfo{author}{Yang, Z.}, \bibinfo{author}{Pan, Z.}, \bibinfo{author}{Li,
  M.}, \bibinfo{author}{Wu, K.}, \bibinfo{author}{Gao, X.},
  \bibinfo{year}{2023}.
\newblock \bibinfo{title}{Learning based 2d irregular shape packing}.
\newblock \bibinfo{journal}{ACM Transactions on Graphics (TOG)}
  \bibinfo{volume}{42}, \bibinfo{pages}{1--16}.
\bibitem[{Zhang et~al.(2022)Zhang, Bai, Qu, Tu and Jin}]{zhang2022deep}
\bibinfo{author}{Zhang, Y.}, \bibinfo{author}{Bai, R.}, \bibinfo{author}{Qu,
  R.}, \bibinfo{author}{Tu, C.}, \bibinfo{author}{Jin, J.},
  \bibinfo{year}{2022}.
\newblock \bibinfo{title}{A deep reinforcement learning based hyper-heuristic
  for combinatorial optimisation with uncertainties}.
\newblock \bibinfo{journal}{European Journal of Operational Research}
  \bibinfo{volume}{300}, \bibinfo{pages}{418--427}.
\bibitem[{Zhao et~al.(2022a)Zhao, Zhu, Xu, Huang and Xu}]{zhao2022learning}
\bibinfo{author}{Zhao, H.}, \bibinfo{author}{Zhu, C.}, \bibinfo{author}{Xu,
  X.}, \bibinfo{author}{Huang, H.}, \bibinfo{author}{Xu, K.},
  \bibinfo{year}{2022}a.
\newblock \bibinfo{title}{Learning practically feasible policies for online 3d
  bin packing}.
\newblock \bibinfo{journal}{Science China Information Sciences}
  \bibinfo{volume}{65}, \bibinfo{pages}{112105}.
\bibitem[{Zhao et~al.(2022b)Zhao, Rao and Fang}]{zhao2022reinforcement}
\bibinfo{author}{Zhao, X.}, \bibinfo{author}{Rao, Y.}, \bibinfo{author}{Fang,
  J.}, \bibinfo{year}{2022}b.
\newblock \bibinfo{title}{A reinforcement learning algorithm for the
  2d-rectangular strip packing problem}, in: \bibinfo{booktitle}{Journal of
  Physics: Conference Series}, \bibinfo{organization}{IOP Publishing}. p.
  \bibinfo{pages}{012002}.
\bibitem[{Zhu et~al.(2020)Zhu, Ji and Li}]{zhu2020hybrid}
\bibinfo{author}{Zhu, K.}, \bibinfo{author}{Ji, N.}, \bibinfo{author}{Li,
  X.D.}, \bibinfo{year}{2020}.
\newblock \bibinfo{title}{Hybrid heuristic algorithm based on improved rules \&
  reinforcement learning for 2d strip packing problem}.
\newblock \bibinfo{journal}{IEEE Access} \bibinfo{volume}{8},
  \bibinfo{pages}{226784--226796}.
\bibitem[{Zuo et~al.(2022)Zuo, Liu, Xu, Xiao, Xu, Liu and Chan}]{zuo2022three}
\bibinfo{author}{Zuo, Q.}, \bibinfo{author}{Liu, X.}, \bibinfo{author}{Xu, L.},
  \bibinfo{author}{Xiao, L.}, \bibinfo{author}{Xu, C.}, \bibinfo{author}{Liu,
  J.}, \bibinfo{author}{Chan, W.K.V.}, \bibinfo{year}{2022}.
\newblock \bibinfo{title}{The three-dimensional bin packing problem for
  deformable items}, in: \bibinfo{booktitle}{2022 IEEE International Conference
  on Industrial Engineering and Engineering Management (IEEM)},
  \bibinfo{organization}{IEEE}. pp. \bibinfo{pages}{0911--0918}.

\end{thebibliography}

\end{document}